\documentclass[11pt]{article}

\usepackage[final]{acl}

\usepackage{times}
\usepackage{latexsym}
\usepackage[T1]{fontenc}
\usepackage[utf8]{inputenc}
\usepackage{microtype}
\usepackage{inconsolata}
\usepackage{graphicx}
\usepackage{amsmath}
\usepackage{amsthm}
\newtheorem{theorem}{Theorem}
\usepackage{hyperref}
\usepackage{amsfonts}
\usepackage{mathtools}
\usepackage{multirow}
\usepackage{booktabs}
\usepackage[table, svgnames]{xcolor}
\definecolor{palegreen}{HTML}{dcf6dd}
\usepackage{tabularx}
\usepackage{algorithm}
\usepackage{algpseudocode}
\usepackage{enumitem}
\newcolumntype{Y}{>{\centering\arraybackslash}X}
\usepackage{graphicx}
\usepackage[most]{tcolorbox}
\definecolor{theorembg}{HTML}{E5E2CA}
\newtcolorbox[auto counter]{theorembox}[2][]{
    colback=theorembg!40,    
    colframe=black!70,  
    boxrule=1.3pt,      
    arc=2mm,            
    left=6pt, right=6pt, top=6pt, bottom=6pt, 
    fonttitle=\bfseries,
    coltitle=black,
    title=Theorem~\thetcbcounter. \textit{#2}., 
    detach title,       
    before upper={\tcbtitle\par\space}, 
    #1                  
}

\title{Stable On-Policy Distillation through Adaptive Target Reformulation}

\author{
Ijun Jang$\quad$ Jewon Yeom$\quad$ Juan Yeo$\quad$  Hyunggyu Lim$\quad$  Taesup Kim$^{\dagger}$ \\
Graduate School of Data Science, Seoul National University \\
\texttt{\{ijun0824, jewon0908, juanyeo, sjrksjek, taesup.kim\}@snu.ac.kr}
}

\begin{document}
\maketitle
\begin{abstract}

Knowledge distillation (KD) is a widely adopted technique for transferring knowledge from large language models to smaller student models. However, conventional supervised KD often suffers from a distribution mismatch between training and inference. While on-policy KD approaches attempt to mitigate this issue by learning directly from student-generated outputs, they frequently encounter training instabilities because the distributional gap between the novice student and the expert teacher is often too wide to bridge directly. These challenges manifest as pathological gradients in forward KL objectives or diversity collapse in reverse KL regimes. To address these limitations, we propose \emph{Veto}, an objective-level reformulation that constructs a geometric bridge in the logit space. Unlike prior methods that mix data samples, \emph{Veto} creates an intermediate target distribution that promotes alignment between the teacher and the student. By introducing a tunable parameter $\beta$, \emph{Veto} serves as an \emph{Adaptive Gradient Veto} that stabilizes optimization by suppressing harmful gradients on low-confidence tokens, while simultaneously acting as a \emph{Decisiveness Knob} to balance reward-driven performance with output diversity. Experiments across reasoning and generation tasks demonstrate that \emph{Veto} consistently outperforms supervised fine-tuning and existing baselines. The code is available at \url{https://github.com/jjun-0824/Veto}.
\end{abstract} 
\renewcommand{\thefootnote}{\fnsymbol{footnote}}
\footnotetext[2]{Corresponding author.}%
\renewcommand{\thefootnote}{\arabic{footnote}} 
\section{Introduction}


\begin{figure*}[t]
  \centering
      \includegraphics[width=0.9\linewidth]{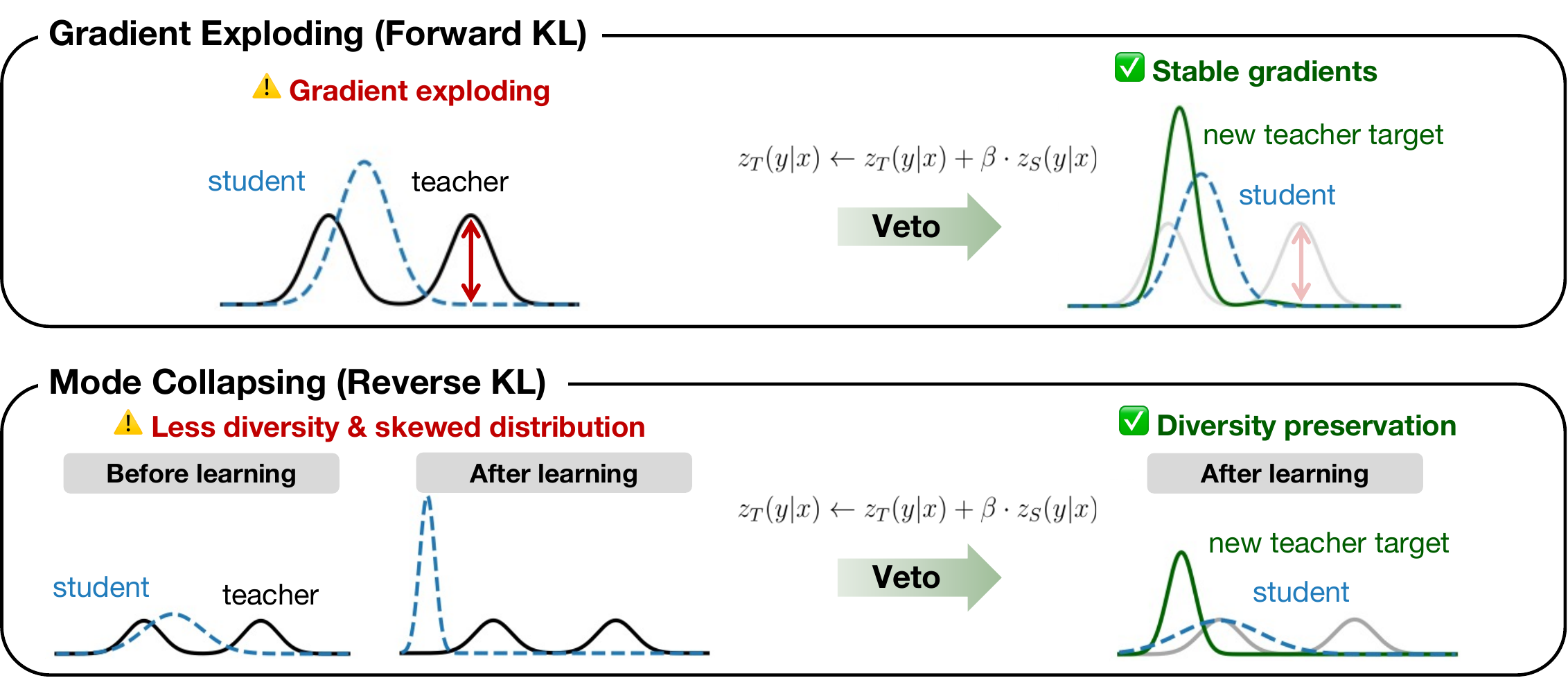}
    \label{fig:veto}
  \caption{\textbf{Overview of stable on-policy knowledge distillation.} \textbf{Top (Forward KL):} Standard Forward KL triggers explosive gradients when students are \emph{ignorant of teacher-preferred tokens} (low $P_S$). Veto reformulates the logit-space target to suppress harmful updates to stabilize early optimization. \textbf{Bottom (Reverse KL)}: Standard Reverse KL causes mode collapse and diversity loss. By tuning $\beta$, Veto balances generation decisiveness with distributional diversity, allowing the student to capture multiple teacher modes effectively.}
\end{figure*}

Knowledge distillation (KD) \cite{kd-2006, kd-2015} is widely used for transferring capabilities from proprietary models to efficient open-source counterparts and facilitating model self-improvement \cite{xu2024survey}. Beyond model compression, KD has recently become an important mechanism for improving large language models (LLMs) through self-training and alignment, where a strong teacher supervises a weaker or partially trained student. However, traditional supervised KD \cite{distill-supervised, distill-seq} suffers from exposure bias \cite{bias-ranzato, bias-bengio}: the mismatch between teacher-provided trajectories and the student’s self-generated outputs leads to degraded performance in autoregressive tasks \cite{autoregressive-zhang, autoregressive-arora}, especially in long-horizon generation and reasoning settings where errors compound over time.

On-policy knowledge distillation addresses this limitation by aligning training with inference-time behavior, learning directly from student-generated outputs \cite{distill-onpolicy, distill-imitkd}. This enables the student to receive feedback on its own predictions and adapt within regions of the probability space it is likely to visit at test time. Building on this idea, recent methods attempt to bridge the teacher-student gap using reverse KL divergences \cite{distill-minillm, distill-fdiv} or interleaved sampling strategies \cite{distill-skd}. While these approaches try to bridge the gap by mixing teacher and student tokens at the \emph{data level}, they largely overlook a critical challenge: \emph{the stability of the optimization objective itself}. Even with mixed data, forcing a novice student to match an expert's sharp distribution creates a steep optimization cliff.

In practice, early-stage student policies are highly noisy and often assign near-zero probability to teacher-preferred tokens. As illustrated in \autoref{fig:output_onpolicy}, this leads to unreliable teacher feedback and pathological gradients. Standard forward KL objectives suffer from gradient explosion on such \emph{ignorant tokens} \cite{distill-onpolicy}, while reverse KL objectives, although numerically stable, lack explicit control over the trade-off between mode-seeking decisiveness and distributional diversity, frequently resulting in premature mode collapse. These failure modes stem from the geometry of the divergence objective, rather than from model architecture or data generation strategy.

In this work, we propose \emph{Veto}, an objective-level reformulation that stabilizes on-policy knowledge distillation by constructing a distributional bridge between the teacher and the student. Instead of mixing data samples, \emph{Veto} mixes the distributions themselves in the logit space to create a geometric target that emphasizes regions of agreement. This intermediate target effectively ``vetoes'' harmful updates on low-confidence tokens, acting as a safe stepping stone for the student. We show that a single scalar parameter $\beta$ plays a dual role: it acts as an \emph{Adaptive Gradient Veto} that prevents gradient explosion in forward KL, and as a \emph{Decisiveness Knob} that enables controlled entropy regularization and explicit diversity-decisiveness trade-offs in reverse KL. Extensive experiments across reasoning and generation tasks demonstrate that \emph{Veto} consistently improves training stability and outperforms supervised fine-tuning and existing on-policy distillation baselines.

\paragraph{Contributions}
Our contributions are summarized as follows:
\begin{itemize}[noitemsep,nosep,leftmargin=1.5em]
    \item We introduce \emph{Veto}, an objective-level reformulation for on-policy knowledge distillation that bridges the teacher-student gap via logit-space interpolation, improving stability without modifying architectures.
    \item We show that a single parameterized target distribution unifies the treatment of forward and reverse KL objectives, mitigating gradient explosion and mode collapse.
    \item We provide theoretical analysis connecting \emph{Veto} to adaptive gradient suppression in forward KL and entropy-regularized policy gradients in reverse KL.
    \item We empirically validate \emph{Veto} on reasoning, code generation, and summarization tasks, demonstrating consistent gains over supervised and on-policy baselines.
\end{itemize}

\section{Related Works}

Knowledge Distillation (KD)~\cite{kd-2006, kd-2015} has become a cornerstone technique for compressing Large Language Models (LLMs) by transferring knowledge from a high-capacity teacher to a more efficient student. Early approaches, often categorized as Supervised KD, typically train the student to mimic the teacher's distribution over a fixed dataset~\cite{distill-supervised} or sequence-level outputs~\cite{distill-seq}. Despite their effectiveness, these methods are prone to exposure bias~\cite{bias-ranzato, bias-bengio}, which arises from the mismatch between ground-truth–conditioned training and self-generated outputs. This discrepancy often results in degraded performance in autoregressive generation tasks~\cite{autoregressive-zhang, autoregressive-chiang, autoregressive-arora}.

To mitigate this training-inference mismatch, recent research has shifted its focus to on-policy KD, which aligns the training distribution with the student's inference distribution. Inspired by imitation learning~\cite{imitation-ross}, ImitKD~\cite{distill-imitkd} initiated this line of work by training the student on a mixture of fixed and self-generated sequences. Building on this, GKD~\cite{distill-onpolicy} introduced a generalized framework that enables distillation entirely on student-generated outputs, demonstrating that learning from self-generated mistakes can substantially improve performance.

Subsequent work has further refined on-policy distillation objectives. MiniLLM ~\cite{distill-minillm} and f-distill~\cite{distill-fdiv} leverage reverse KL and f-divergences to encourage mode-seeking behavior. Furthermore, SKD~\cite{distill-skd} introduces interleaved sampling to actively correct 
low-quality student generations, improving feedback quality. Despite these advances, existing approaches largely overlook the stability of the optimization objective itself, particularly during the early stages of training when student outputs are highly noisy. In this paper, we propose \emph{Veto}, which simultaneously acts as a gradient stabilizer for forward KL and a tunable control for decisiveness in reverse KL, offering a unified solution that ensures stable optimization without the architectural overhead of previous methods.
\begin{figure}[t]
  \centering
    \includegraphics[width=\linewidth]{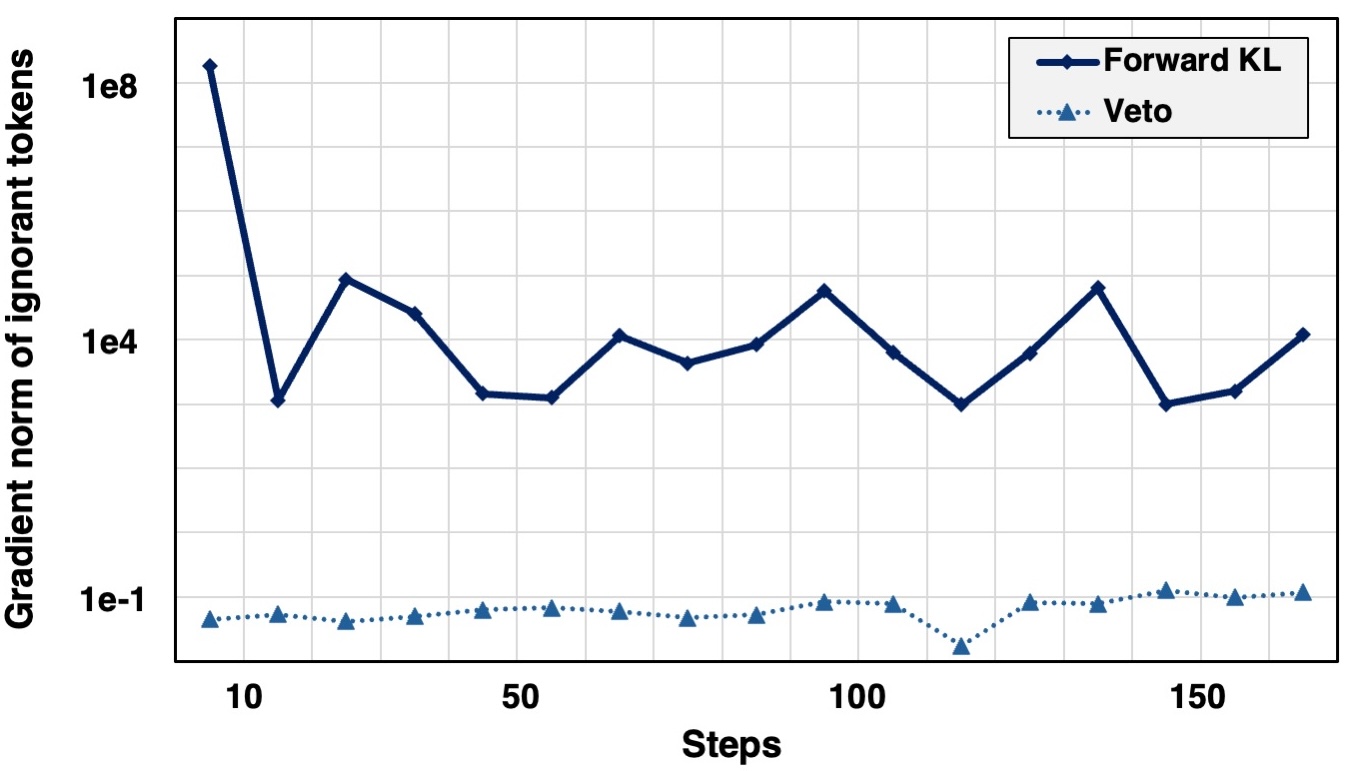}
    \label{fig:gradient}
    \vspace{-1.5em}
  \caption{\textbf{Comparison of gradient magnitudes on 
  ignorant tokens ($P_T > 0.1, P_S < 0.01$).} While standard Forward KL triggers pathological gradient explosions exceeding $10^7$, Veto effectively suppresses these spikes within a stable range. This demonstrates \emph{Veto}’s ability to ensure optimization stability by suppressing harmful updates during early training.}
  \label{fig:gradient}
\end{figure}


\begin{table*}[t!]
\centering
\footnotesize

\begin{tabular}{p{0.97\linewidth}}
\toprule
\texttt{<Example 1>}\\
\textbf{Question:} There are 350 trees in the park, 70\% of which are pine trees. How many are not pine trees?\\
\midrule

\textbf{Answer:} To find out how many trees are not pine trees, we need to subtract the number of pine trees from the total number of trees. $\rightarrow$ \texttt{\textcolor{red}{NO ANSWER}}
\\
\midrule
\textbf{Reference:} There are 350 x 70 / 100 = <<350*70/100=245>> 245 pine trees. Therefore, 350 - 245 = <<350-245=105>> 105 trees are not pine trees. $\rightarrow$ \texttt{\textcolor{blue}{105}}
\\
\midrule\midrule
\texttt{<Example 2>}\\
\textbf{Question:} The Donaldsons pay \$15 per hour for babysitting. The Merck family pays \$18 per hour and the Hille family pays \$20 per hour for babysitting. Layla babysat for the Donaldsons for 7 hours, the Merck family for 6 hours and the Hille family for 3 hours. How many dollars did Layla earn babysitting? 
\\
\midrule
\textbf{Answer:} To find out how much Layla earned in total, we need to multiply the number of hours she worked by her hourly rate. $\rightarrow$ \texttt{\textcolor{red}{NO ANSWER}}
\\
\midrule
\textbf{Reference:} Donaldsons pay \$15 / hour = 15 * 7 = <<15*7=105>> 105. The Mercks pay \$18 / hour = 18 * 6 = <<18*6=108>> 108. The Hilles pay \$20 / hour = 20 * 3 = <<20*3=60>> 60. The total amount Layla earned is 105 + 108 + 60 = <<105+108+60=273>> 273. Layla earned \$273 for babysitting. $\rightarrow$ \texttt{\textcolor{blue}{273}}\\
\midrule\midrule
\texttt{<Example 3>}\\
\textbf{Question:} Bryan works as a social media account manager. He does marketing posts, advertisement posts, and customer outreach posts. His client has him spend four hours on customer outreach posts and half that time on advertisement posts each day. Bryan works eight hours a day. How much time in hours each day does he spend on marketing posts? 
\\
\midrule

\textbf{Answer:} He spends 4/2 = 2 hours on advertisement posts.
So he spends 8-4-2 = 0 hours on marketing posts.\\ $\rightarrow$ \texttt{\textcolor{red}{WRONG LOGIC}}
\\
\midrule
\textbf{Reference:} Let M be the number of hours Bryan spends on marketing posts. He spends 4 / 2 = <<4/2=2>> 2 hours on advertisement posts a day. He works M + 4 + 2 = 8 hours a day. Thus, Bryan spends \textcolor{blue}{ M = 8 - 4 - 2 = <<8-4-2=2>> 2} hours on marketing posts each day. $\rightarrow$ \texttt{\textcolor{blue}{2}}
\\
\midrule\midrule
\texttt{<Example 4>}\\
\textbf{Question:} Drew is reseeding his lawn with grass seed. One bag of grass seed covers 250 square feet of lawn. His lawn is 22 feet from the house to the curb and 36 feet from side to side. He bought four bags of seed. How many extra square feet could the leftover grass seed cover after Drew reseeds his lawn?
\\
\midrule

\textbf{Answer:} The area of the lawn is 22*36 = 792 square feet. Four bags of grass seed cover 4*250 = 1000 square feet. So, the leftover grass seed can cover {\textcolor{red}{792-1000 = -208 square feet}}. Since the result is negative, it means that the leftover grass seed cannot cover any more area. $\rightarrow$ \texttt{\textcolor{red}{WRONG LOGIC}}
\\
\midrule
\textbf{Reference:} Drew \u2019s lawn is 22 * 36 = <<22*36=792>> 792 square feet. He bought enough grass seed to cover 250 * 4 = <<250*4=1000>> 1000 square feet. Thus, the leftover grass seed can cover 1000 - 792 = <<1000-792=208>> 208 square feet.
$\rightarrow$ \texttt{\textcolor{blue}{208}}
\\
\bottomrule
\end{tabular}
  \caption{\textbf{Low-quality outputs of On-policy KD.} Student produces low-quality samples, leading to unreliable or noisy teacher feedback. Optimization is hindered by pathological gradients stemming from noisy teacher feedback on low-quality, out-of-distribution student samples.}    \label{fig:output_onpolicy}
\end{table*}

\section{Preliminaries}

\subsection{On-Policy Knowledge Distillation}
Knowledge distillation (KD) \cite{kd-2006, kd-2015} aims to transfer the predictive capabilities of a high-capacity teacher model $P_T$ to a smaller student model $P_S$. In the context of Large Language Models (LLMs), conventional supervised KD typically minimizes the divergence between the student's output and fixed teacher-generated trajectories. However, this approach often leads to exposure bias \cite{bias-ranzato}, as the student is not trained to handle the distribution shifts it encounters during autoregressive generation at inference time.

To mitigate this, on-policy KD \cite{distill-onpolicy} aligns the training distribution with the student's inference distribution by sampling sequences $y = (y_1, \dots, y_T)$ directly from the student's own policy $P_S(\cdot \mid x)$. Formally, for a given prompt $x$, the objective is to minimize the expected token-level divergence along these self-generated trajectories:
\begin{equation*}
\begin{aligned}
\mathcal{L} &= \mathbb{E}_{x \sim P_{\text{data}}, y \sim P_S(\cdot \mid x)} \\
&\quad \left[ \sum_{t=1}^{|y|} D_{\text{KL}}(P_T(\cdot \mid x, y_{<t}) \,\|\, P_S(\cdot \mid x, y_{<t})) \right]
\end{aligned}
\end{equation*}
By receiving feedback on its own mistakes, the student learns to correct its behavior within the specific regions of the probability space it is likely to visit during test time \cite{distill-onpolicy, distill-skd}.

\subsection{Asymmetric Properties of KL Divergence}
The behavior of on-policy distillation is fundamentally governed by the direction of the Kullback-Leibler (KL) divergence. The forward KL divergence, $D_{\text{KL}}(P_T \,\|\, P_S)$, is known for its \emph{zero-avoiding} property, which forces the student to assign non-zero probability to any token the teacher considers likely. In contrast, the reverse KL divergence, $D_{\text{KL}}(P_S \,\|\, P_T)$, is \emph{mode-seeking}, encouraging the student to concentrate its probability mass on the primary modes of the teacher's distribution. While these properties provide the theoretical basis for various distillation frameworks \cite{distill-minillm, distill-fdiv}, they also introduce significant numerical challenges when applied to on-policy settings where the student’s early-stage outputs are highly unstable.

\begin{algorithm}[t]
\caption{Stable On-policy Distillation via \emph{Veto} with Linear $\beta$ Decay}
\label{alg:veto_linear}
\begin{algorithmic}[1]
\Require Student $P_{\theta}$, Teacher $P_T$, Dataset $\mathcal{X}$, Initial hyperparameter $\beta$, Total steps $N$, Learning rate $\eta$
\State \textbf{Initialize} student parameters $\theta$
\For{each training step $i = 1 \dots N$}
    \Statex \hspace{1em}\textbf{$\vartriangleright$ Linear decay from $\beta$ to 0}
    \State $\beta \leftarrow \beta \cdot \left( 1 - \frac{i}{N} \right)$    
    \Statex \hspace{1em}\textbf{$\vartriangleright$ On-policy sampling}
    \State Sample prompt $x \sim \mathcal{X}$
    \State Sample trajectory $y \sim P_{\theta}(\cdot|x)$
    \For{each token $t=1 \dots |y|$}
        \Statex \hspace{2.5em} \textbf{$\vartriangleright$ Get logits}
        \State $z_T = \log P_T(\cdot|x, y_{<t})$
        \State $z_S = \log P_{\theta}(\cdot|x, y_{<t})$ 
        \Statex \hspace{2.5em} \textbf{$\vartriangleright$ Compute reformulated target}
        \State $Q \propto \exp(z_T + \beta \cdot z_S)$
        \If{using Forward KL regime}
            \Statex \hspace{4em} \textbf{$\vartriangleright$ Adaptive gradient \emph{Veto}ing}
            \State $\mathcal{L}_t = D_{KL}(Q \parallel P_{\theta})$ 
        \ElsIf{using Reverse KL regime}
            \Statex \hspace{4em} \textbf{$\vartriangleright$ Decisiveness control}
            \State $\mathcal{L}_t = D_{KL}(P_{\theta} \parallel Q)$ 
        \EndIf
    \EndFor    
    \Statex \hspace{1em} \textbf{$\vartriangleright$ Optimization}
    \State $\mathcal{L} = \sum_{t} \mathcal{L}_t$
    \State $\theta \leftarrow \theta - \eta \nabla_{\theta} \mathcal{L}$ 
\EndFor
\end{algorithmic}
\end{algorithm}

\section{Methodology}

\subsection{Stability Challenges in On-Policy KD}
Despite the theoretical advantages of learning on-policy, the mismatch between the teacher's expertise and the student's initial ignorance often leads to optimization failures.

\paragraph{Gradient Explosion in Forward KL} 
The forward KL gradient with respect to student parameters $\theta$ is defined as $\nabla_\theta \mathcal{L}_{\mathrm{FWD}} \approx - \sum_y \frac{P_T(y)}{P_S(y)} \nabla_\theta P_S(y)$. When the student is ignorant of teacher-preferred tokens ($P_S(y) \to 0$ while $P_T(y) > 0$), the ratio $\frac{P_T(y)}{P_S(y)}$ diverges. As shown in \autoref{fig:gradient}, this leads to pathological gradient explosions during early training stages, where student is ignorant of teacher-preferred tokens \cite{distill-onpolicy}.

\paragraph{Mode Collapse in Reverse KL} 
While reverse KL avoids divergence, it lacks a mechanism to explicitly control the intensity of mode-seeking behavior. 
As a result, reverse-KL based distillation can encourage the student to concentrate on dominant teacher modes, potentially reducing distributional diversity.

\subsection{\emph{Veto}: Bridging the Gap via Target Reformulation}

Standard on-policy KD often suffers from a severe distribution mismatch: forcing an early-stage student to immediately match an expert teacher's complex distribution can lead to optimization instability. To address this, we propose \emph{Veto}, an objective-level reformulation that acts as a \textbf{geometric bridge} between the teacher and the student.

Instead of targeting the teacher directly, we construct an intermediate target distribution $Q$ in the logit space. Let $z_T = \log P_T(\cdot \mid x)$ and $z_S = \log P_S(\cdot \mid x)$ denote the logits. We define $Q$ as:
\vspace{-0.5em}
\begin{equation*}
\begin{aligned}
Q(y \mid x) &\propto \exp(z_T(y \mid x) + \beta \cdot z_S(y \mid x)) \\
&= \frac{1}{Z(x)} P_T(y \mid x) \cdot P_S(y \mid x)^{\beta}
\end{aligned}
\end{equation*}
In this formulation, Q defines a \textbf{geometric bridge} by forming an intermediate distribution that geometrically interpolates the teacher and student distributions in probability space. This formulation represents a \emph{Product of Experts (PoE)}. Intuitively, it works as a \textbf{consensus filter}: the target probability $Q(y|x)$ is high only when the token is supported by \emph{both} the teacher (quality) and the student (confidence). The parameter $\beta \geq 0$ controls the position of this bridge, allowing the student's current ignorance to effectively \emph{Veto} overwhelming signals from the teacher.

\begin{table*}[t!]
  \centering
  \small
  \begin{tabularx}{\textwidth}{l *{4}{Y}} 
    \toprule
    \multirow{2.5}{*}{\textbf{Method}} & \bf GSM8K & \multicolumn{2}{c}{\bf HumanEval} & \bf DialogSum \\ 
    \cmidrule(lr){2-2} \cmidrule(lr){3-4} \cmidrule(lr){5-5}
     & \bf Accuracy & \bf Pass@1 & \bf Pass@10 & \bf Win-rate \\
    \midrule
    Teacher SFT   & 74.7 & 64.7 & 72.2 & 65.0 \\
    \midrule
    Student SFT   & 30.7 & 26.9 & 34.6  & 54.0 \\
    Supervised KD & 33.4 & 26.8 & 34.5  & 54.3 \\
    SKD           & 33.6 & 24.8 & 34.8  & 53.6 \\
    On-policy KD  & 35.1 & 22.9 & 35.3 & 54.3 \\
    \rowcolor{PaleGreen!30} 
    \textbf{Ours (\emph{Veto})} & \textbf{39.9} & \textbf{29.0} & \textbf{37.7} & \textbf{56.5} \\
    \bottomrule
  \end{tabularx}
  \caption{\textbf{Performance across three distinct domains.} We evaluate mathematical reasoning on GSM8K (Accuracy), code generation on HumanEval (Pass@1, Pass@10), and dialogue summarization on DialogSum (Win-rate). Pass@1 and Pass@10 quantify the proportion of problems for which at least one correct solution is obtained from 1 and 10 generated samples, respectively, whereas the Win-rate represents the proportion of model outputs that are preferred by GPT-4o-mini in pairwise evaluations. Our method outperforms baselines in all student categories. Note that we use $\beta=0.8$ for reasoning, $\beta=1.0$ for code tasks, and $\beta=0.3$ for summarization.}  
  \label{tab:main}
\end{table*}


\subsection{Theoretical Analysis}
We provide a unified analysis of \emph{Veto}, explicitly showing how the reformulated target ensures stability (Analysis I) and bridges KD with RL (Analysis II). Detailed proofs are in Appendix~\ref{sec:appendix_proofs}.

\paragraph{Analysis I: Stability via Adaptive \emph{Veto}}
In standard Forward KD, the loss explodes when the student assigns near-zero ($P_S \to 0$) probability to a token the teacher prefers, as the term $\log P_S$ diverges. Veto resolves this structurally.

\vspace{1.0em}

\begin{theorem}[Adaptive Gradient \emph{Veto}]
In standard Forward KD ($\beta = 0$), gradients diverge as $P_S(y) \to 0$. However, in Veto ($\beta > 0$), the target $Q$ incorporates the student's uncertainty, ensuring the loss converges to 0.
\\
\end{theorem}
\noindent \emph{Proof Intuition.}
As shown in Appendix~\ref{sec:appendix_proofs}, substituting $Q$ into the loss
gradient, the diverging term becomes proportional to
\begin{equation*}
\mathcal{L}(y) \approx  P_S(y)^{\beta} \log P_S(y).
\end{equation*}
By L'Hôpital's rule, the polynomial term $P_S^\beta(y)$ decays to zero faster than
the logarithmic term $\log P_S(y)$ diverges, effectively acting as a gate that suppresses
updates on ignorant tokens.
\vspace{1.0em}

\begin{theorem}[Sharpening Effect]
The student converges to a sharpened version of the teacher with implicit temperature scaling $T = 1 - \beta$.
\end{theorem}
\noindent \emph{Derivation Logic.} At the optimal fixed point where $P_S^* = Q$, we have the relation $P_S^* \propto P_T \cdot (P_S^*)^\beta$. Rearranging terms yields $(P_S^*)^{1-\beta} \propto P_T$, which implies:
\begin{equation*}
P_S^*(y \mid x) \propto P_T(y \mid x)^{\frac{1}{1-\beta}}
\end{equation*}
Since $0 \le \beta < 1$, the exponent $\frac{1}{1-\beta} > 1$, naturally encouraging the student to be more decisive (sharper) than the teacher.

\paragraph{Analysis II: Bridge to Reinforcement Learning}
In the reverse regime, \emph{Veto} acts as a \emph{Decisiveness Knob} by bridging the gap between KD and RL.

\vspace{1.0em}


\begin{theorem}[Bridge to REINFORCE]
The gradient of the Reverse GKD objective with respect to the student
parameters $\theta$ is equivalent to a policy-gradient update with scaled
entropy regularization:
\begin{equation*}
\nabla_\theta \mathcal{L}_{\mathrm{REV}}
=
\mathbb{E}_{y \sim P_S}
\Big[
\nabla_\theta \log P_S(y)
\cdot A(y)
\Big],
\end{equation*}
where
\begin{equation*}
A(y)
=
\underbrace{-\log P_T(y)}_{\text{Reward } r(y)}
\;+\;
\underbrace{(1-\beta)\log P_S(y)}_{\text{Scaled Entropy Cost}}.
\end{equation*}
\end{theorem}

\vspace{1.0em}

\paragraph{The Spectrum of Decisiveness} This formulation reveals that $\beta$ controls the trade-off between mode-covering and mode-seeking:
\begin{itemize}
\item \textbf{$\beta = 0$ (Standard Reverse KD):} The student tries to match the teacher's distribution exactly, inheriting both its modes and its uncertainty.
\item \textbf{$0 < \beta < 1$ (Geometric KD):} The "Middle Ground." The student seeks high-reward regions (like RL) while retaining a diversity budget proportional to $(1-\beta)$ to prevent premature collapse.
\item \textbf{$\beta \to 1$ (Pure RL):} Zero entropy regularization. The objective becomes equivalent to \textsc{REINFORCE}, where the student collapses to the single highest-reward mode.
\end{itemize}


\section{Experiments}

\subsection{Implementation Details}

\paragraph{Model details} We use Qwen2-0.5B-IT~\cite{model-qwen2} as the student model and Qwen2-7B-IT~\cite{model-qwen2} as the teacher model. 
We first fine-tune the teacher model in a supervised manner on task-specific data. The student model is then trained using outputs generated via on-policy KD, and optimized using our proposed objective with forward KL. 

\paragraph{Hyperparameters} All experiments use a learning rate of 1e-5, a warmup ratio of 0.1, and a dropout rate of 0.1, with training conducted for three epochs with 2 H100 GPUs with 80GB of memory. The hyperparameter $\beta$ was selected via grid search and scheduled with linear decay (see \nameref{para:scheduling}).

\subsection{Mathematical Reasoning}
We conduct experiments on the GSM8K~\cite{data-gsm8k} dataset. Following the setup used in the ~\cite{distill-skd}, we first fine-tune the teacher model in a supervised manner on 7K instances from the GSM8K training set. We then randomly sample 1K instances from the training set to construct our input–output pairs $(x, y)$ for student training. For evaluation, we employ the GSM8K test set, which consists of 1,319 instances. The quality of generated outputs is assessed using answer accuracy.

As shown in \autoref{tab:main}, our method significantly outperforms supervised fine-tuning of the student model, yielding a 9.2$\%$ absolute improvement in accuracy (30.7$\%$ → 39.9$\%$). Moreover, our approach surpasses prior baselines, including SKD and on-policy KD, by up to 6.3$\%$, demonstrating the effectiveness of our objective-level reformulation in on-policy distillation settings.

\subsection{Code Generation}
For code generation, we use prompts from the WizardCoder~\cite{data-wizardcoder} dataset. From this dataset, we randomly sample 10K instances for supervised fine-tuning of the teacher model, 1K instances for student training, and 1K instances for evaluation.

We evaluate code generation performance on the HumanEval~\cite{eval-humaneval} benchmark using pass@k metrics. 
As shown in \autoref{tab:main}, compared to on-policy KD, our method improves Pass@1 from 22.9 to 29.0 (+6.1) and Pass@10 from 35.3 to 37.7 (+2.4), indicating clear gains in code synthesis performance.

\subsection{Summarization}
For dialogue summarization, we utilize the DialogSum~\cite{data-dialogsum} dataset. We randomly sample 1K instances from the DialogSum training set to construct input–output pairs $(x, y)$ for student training. For evaluation, we use the official test set (1,500 instances). We first fine-tune the teacher model in a supervised manner on 10K instances sampled from the DialogSum training set, and then train the student model on the 1K sampled instances using our proposed distillation method. Summary quality is evaluated using win-rate~\cite{eval-winrate} based on pairwise comparisons judged by GPT-4o-mini. 

As reported in \autoref{tab:main}, our method achieves the highest win-rate among all student models. 
In particular, it improves the win-rate from 54.3 to 56.5 (+2.2) compared to both supervised KD and on-policy KD, and outperforms SKD by 2.9 $\%$ (53.6$\%$ → 56.5$\%$). 
These results indicate that our approach produces more faithful and higher-quality summaries, even when trained with limited student data.

\vspace{0.5em}

\subsection{Ablation Study}
For ablation study, we tested on GSM8k with Qwen2-0.5B-IT~\cite{model-qwen2} as the student model and Qwen2-7B-IT~\cite{model-qwen2} as the teacher model. Unless otherwise specified, the default objective for our experiments is conducted on on-policy outputs with \emph{Veto} (forward KL) objective.

\begin{figure}[t]
    \centering
    \includegraphics[width=\linewidth]{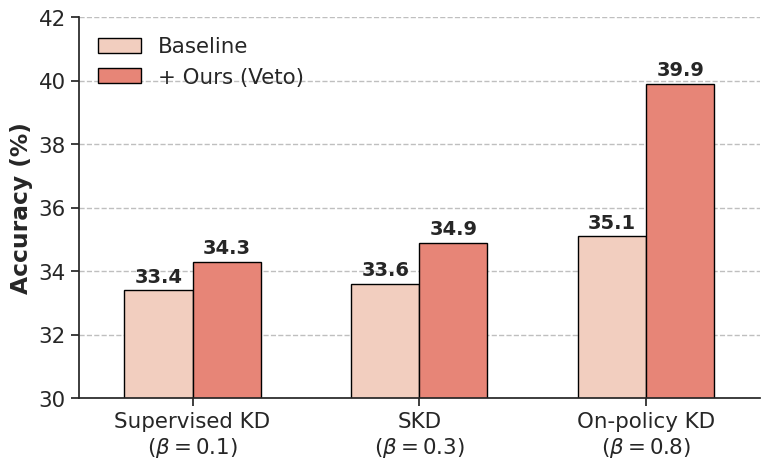}
    \caption{\textbf{Effect of different on-policy data generation strategies} on student performance for mathematical reasoning. Accuracy (\%) is reported along with absolute improvements over the corresponding supervised student baseline for supervised KD, SKD, and on-policy KD under different values of $\beta$.}
    \label{fig:ablation_data}
\end{figure}

\paragraph{Data Generation Setting} In \autoref{fig:ablation_data}, we evaluate the effectiveness of our proposed method across various data generation strategies commonly used in knowledge distillation. Specifically, we apply our objective to supervised KD, speculative KD (SKD), and on-policy KD to assess its robustness. For supervised KD, which relies on a fixed teacher-generated dataset, our method achieves an accuracy of 34.3\%, representing an absolute improvement of 0.9\% over the standard baseline. Similarly, in the SKD setting, we observe a 1.3\% performance gain. Most notably, the most significant improvement occurs in the on-policy setting, where our method reaches 39.9\% accuracy, a +4.8\% gain compared to previous on-policy distillation. These results demonstrate that our proposed objective consistently improves student performance and ensures optimization stability regardless of the specific data generation strategy utilized.

\begin{figure}[t]
    \centering
    \includegraphics[width=0.8\linewidth]{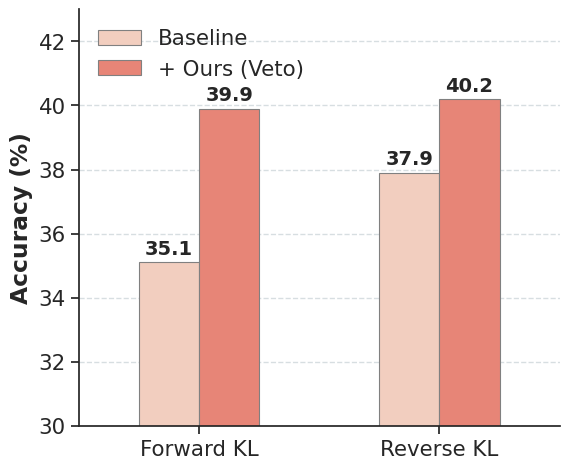}
    \caption{\textbf{Ablation comparing forward and reverse KL divergence objectives.} Accuracy (\%) is reported for standard on-policy knowledge distillation using each KL loss and for the same settings augmented with the proposed method.}
    \label{fig:ablation_loss}
\end{figure}

\paragraph{Adaptability of KL Loss} 
We conduct an ablation study to evaluate the generalizability of \emph{Veto} across different KL divergence objectives: Forward KL and Reverse KL. As summarized in~\autoref{fig:ablation_loss}, our proposed method consistently enhances performance regardless of the underlying divergence used for distillation. By applying \emph{Veto} to forward KL, we observe a significant absolute accuracy gain of 4.8\% (35.1\% $\rightarrow$ 39.9\%). This improvement validates \emph{Veto}'s role as an \emph{Adaptive Gradient Veto}, which stabilizes early-stage optimization by effectively suppressing pathological gradients on ignorant tokens. When applied to Reverse KL, our method also improves performance by 2.3\% (37.9\% $\rightarrow$ 40.2\%). In this context, \emph{Veto} acts as a \emph{Decisiveness Knob}, allowing for controlled entropy regularization and preventing the student from converging too prematurely to a single mode. These results demonstrate that \emph{Veto} is objective-agnostic. It serves as a versatile, objective-level reformulation that provides a unified solution for ensuring optimization stability and enhancing generation quality in various on-policy distillation settings.

\begin{figure}[t]
    \centering
    \includegraphics[width=\linewidth]{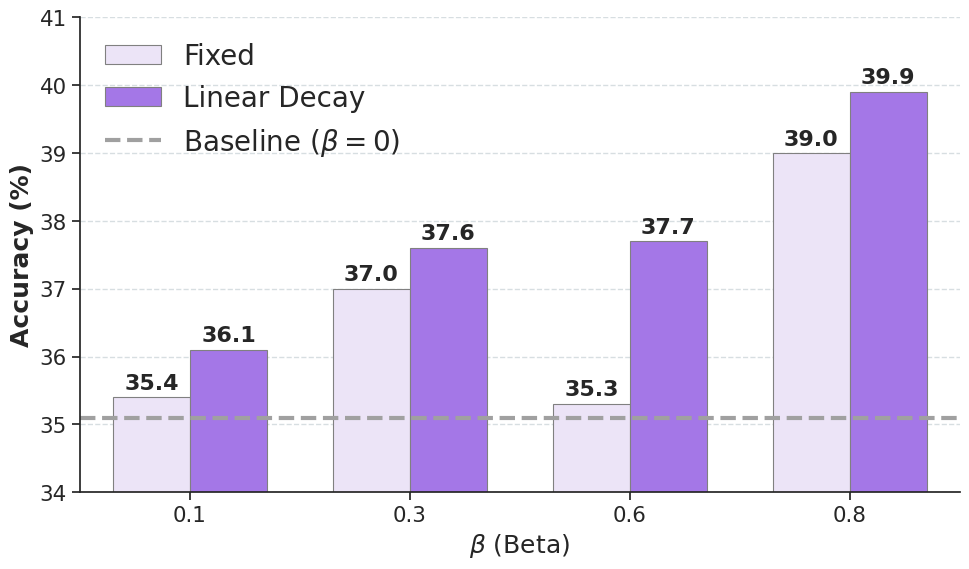}
    \caption{\textbf{Ablation on the $\beta$ scheduling strategy.} Accuracy (\%) of the fixed schedule and the linear decay schedule across different initial $\beta$ values. The dashed line denotes the baseline ($\beta = 0$). The linear decay schedule consistently outperforms the fixed schedule, with an initial $\beta = 0.8$ achieving the best performance.}
    \label{fig:ablation_beta}
\end{figure}

\paragraph{Choice of $\boldsymbol{\beta}$ Scheduling Strategy.} 
\label{para:scheduling}
To further investigate the dynamics of the parameter $\beta$, we explore various scheduling strategies. In our framework, a higher $\beta$ prioritizes optimization stability by more aggressively suppressing pathological gradients on tokens where the student is ignorant. Conversely, a lower $\beta$ allows the student to follow the teacher’s distribution more closely, which is beneficial once the student has gained sufficient proficiency. Based on this intuition, we implemented a linear decreasing schedule to ensure stability during the critical early stages of training. In this regime, $\beta$ starts at a high initial value and linearly decays to 0 as the student's distribution gradually aligns with that of the teacher. We compared this strategy with a constant schedule, where $\beta$ remains fixed throughout the entire training process. As shown in \autoref{fig:ablation_beta}, the linear decreasing schedule consistently outperforms the constant scheduling across all tested values. Specifically, we found that starting from an initial $\beta = 0.8$ yields the highest performance. This suggests that while a strong \emph{Veto} gate is essential for stabilizing the early ignorant phase of the student, gradually relaxing this constraint allows the model to capture the finer nuances of the teacher's distribution as training progresses.


\begin{table}[t]
  \centering
  \resizebox{0.7\linewidth}{!}{
    \begin{tabular}{lcc}
      \toprule
      \textbf{Method} & \textbf{$\beta$} & \textbf{Accuracy} \\
      \midrule
      Teacher SFT & - & 77.2 \\
      \midrule
      Student SFT & - & 49.3 \\
      Supervised KD & - & 47.6 \\
      SKD & - & 47.1 \\
      On-policy KD & - & 49.8 \\
      \rowcolor{PaleGreen!30}
      \textbf{Ours (\emph{Veto})} & \textbf{0.3} & \textbf{51.2} \\
      \bottomrule
    \end{tabular}
  }
  \caption{Mathematical reasoning accuracy on \textbf{GSM8K}. We distill Gemma2-2B-IT using SFTed Gemma2-9B-IT as a teacher. \textbf{Ours (Veto)} shows the best performance among student models.}
  \label{tab:gemma2}
\end{table}

\paragraph{Generalizability to Various Models} To evaluate the generalizability of our proposed method, we extend our experiments to the Gemma2 model family~\cite{model-gemma2}, employing Gemma2-9B-IT as the teacher and Gemma2-2B-IT as the student. We conducted evaluations on the GSM8K dataset, a representative benchmark for mathematical reasoning, with the decisiveness parameter fixed at $\beta = 0.3$. Our results show that \emph{Veto} consistently yields significant performance improvements in this setting as well. This confirms that our objective-level reformulation is not limited to a specific model architecture but is robustly applicable across various state-of-the-art model families and complex reasoning tasks.
\vspace{1.0em}

\section{Conclusion}

In this work, we proposed Veto, an objective-level reformulation for on-policy knowledge distillation (KD) that improves optimization stability. By constructing a geometric target distribution in logit space, Veto emphasizes agreement between the teacher and the student, effectively suppressing the pathological gradients typically encountered in forward KL objectives. Simultaneously, it serves as a Decisiveness Knob in reverse KL regimes, balancing reward-driven performance with distributional diversity to prevent premature mode collapse. Experiments across reasoning, code generation, and summarization tasks demonstrate that Veto consistently outperforms supervised fine-tuning and existing on-policy distillation baselines. These results suggest that Veto is an objective-agnostic and versatile framework capable of enhancing the efficiency and generation quality of smaller language models across diverse domains.


\clearpage
\section*{Acknowledgment}
This work was supported by the National Research Foundation of Korea (NRF) grant funded by the Korea government (MSIT) (No. RS-2024-00345809, Research on AI Robustness Against Distribution Shift in Real-World Scenarios; and No. RS-2023-00222663, Center for Optimizing Hyperscale AI Models and Platforms).
\bibliography{custom}

\clearpage

\appendix
\section{Detailed Mathematical Proofs}
\label{sec:appendix_proofs}

\subsection{Proof of Theorem 1 (Adaptive Gradient Veto)}
In standard forward KD ($\beta=0$), the loss is defined as $\mathcal{L} \approx -c \log P_S(y)$. As $P_S(y) \to 0$, $\log P_S(y) \to -\infty$, causing the loss and its gradient to diverge. 

In Veto ($\beta > 0$), substituting the PoE target $Q \propto P_T P_S^\beta$ into the loss gives:
\begin{equation}
\mathcal{L} \approx -\tilde{c} P_S(y)^\beta \log P_S(y)
\end{equation}
To evaluate the limit as $P_S(y) \to 0^+$, let $p = P_S(y)$. Applying L'Hôpital's rule to the form $\frac{\log p}{p^{-\beta}}$:
\begin{equation}
\lim_{p \to 0^+} \frac{\log p}{p^{-\beta}} = \lim_{p \to 0^+} \frac{1/p}{-\beta p^{-\beta-1}} = \lim_{p \to 0^+} -\frac{p^\beta}{\beta} = 0
\end{equation}
Thus, the loss converges to 0, and the term $P_S(y)^\beta$ effectively ``vetoes'' the updates for tokens where the student is ignorant.

\subsection{Proof of Theorem 2 (Sharpening Effect)}
The forward KL divergence $D_{\text{KL}}(Q \| P_S)$ is minimized when the student distribution $P_S$ matches the target $Q$. Substituting the definition of $Q$ from Eq. (4):
\begin{equation}
\begin{aligned}
P_S^*(y \mid x) &= Q(y \mid x) \\
P_S^* &\propto P_T \cdot (P_S^*)^\beta \\
(P_S^*)^{1-\beta} &\propto P_T \\
P_S^* &\propto P_T^{\frac{1}{1-\beta}}
\end{aligned}
\end{equation}
For $0 < \beta < 1$, the exponent $\gamma = \frac{1}{1-\beta}$ is greater than 1. This confirms that the student converges to a sharpened version of the teacher distribution.

\subsection{Proof of Theorem 3 (Bridge to REINFORCE)}
The reverse KL objective is $J(\theta) = D_{\text{KL}}(P_S \| Q)$. Treating $Q$ as a fixed target for the gradient step:
\begin{equation}
\begin{aligned}
\nabla J(\theta) &= \nabla \sum_y P_S(y) \left( \log P_S(y) - \log Q(y) \right) \\
&= \sum_y \nabla P_S(y) \left( \log P_S(y) - \log Q(y) \right) \\
&\quad + \sum_y P_S(y) \nabla \log P_S(y) \\
&= \mathbb{E}_{y \sim P_S} [ \nabla \log P_S(y) (\log P_S(y) \\
&\quad - (\log P_T(y) + \beta \log P_S(y))) ] \\
&= \mathbb{E}_{y \sim P_S} [ \nabla \log P_S(y) \\
&\quad \cdot ((1-\beta) \log P_S(y) - \log P_T(y)) ]
\end{aligned}
\end{equation}
The term $(1-\beta) \log P_S(y)$ acts as a scaled entropy cost, while $\log P_T(y)$ serves as the reward signal.


\end{document}